\documentclass{article}

\PassOptionsToPackage{numbers, sort&compress}{natbib}
\usepackage[final]{neurips_2023}

\usepackage[utf8]{inputenc} 
\usepackage[T1]{fontenc}    
\usepackage{hyperref}       
\usepackage{url}            
\usepackage{booktabs}       
\usepackage{amsfonts}       
\usepackage{nicefrac}       
\usepackage{microtype}      
\usepackage{xcolor}         
\usepackage{dirtytalk}
\usepackage{graphicx,wrapfig}

\usepackage{amsmath}
\usepackage{amsfonts}
\usepackage{dsfont}
\usepackage{amssymb}
\usepackage{amsthm}
\usepackage{dsfont}
\usepackage{xcolor}
\usepackage{url}
\usepackage{comment}
\usepackage{natbib}
\usepackage{float}
\usepackage{caption}
\usepackage{subcaption}
\usepackage{graphicx}
\usepackage{mathabx}
\usepackage{ulem}
\usepackage{natbib}

\newtheorem*{theorem*}{Theorem}
\newtheorem*{proposition*}{Proposition}

\newtheorem{lemma*}{Lemma}

\newtheorem*{corollary*}{Corollary}

\newtheorem*{definition*}{Definition}

\theoremstyle{remark}
\newtheorem{remark}{Remark}

\newcommand{\R}{\mathbb{R}}
\newcommand{\C}{\mathbb{C}}
\DeclareMathOperator*{\E}{\mathbb{E}}
\newcommand{\cF}{\mathcal{F}}
\newcommand{\cH}{\mathcal{H}}
\newcommand{\kspatial}{K_{\rm spatial}}
\newcommand{\kspectral}{K_{\rm spectral}}

\title{DFU: scale-robust diffusion model for zero-shot super-resolution image generation}

\author{%
  Alex Havrilla \\
  Department of Mathematical Sciences\\
  Georgia Institute of Technology\\
  Atlanta, GA 30332 \\
  \texttt{ahavrilla3@gatech.edu} \\
  \And
  Kevin Rojas \\
  Department of Mathematical Sciences\\
  Georgia Institute of Technology\\
  Atlanta, GA 30332 \\
  \texttt{kevin.rojas@gatech.edu} \\
  \And
  Wenjing Liao \\
  Department of Mathematical Sciences\\
  Georgia Institute of Technology\\
  Atlanta, GA 30332 \\
  \texttt{wliao60@gatech.edu} \\
  \And
  Molei Tao \\
  Department of Mathematical Sciences\\
  Georgia Institute of Technology\\
  Atlanta, GA 30332 \\
  \texttt{mtao@gatech.edu} \\
}

\begin{document}

\maketitle

\begin{abstract}
    Diffusion generative models have achieved remarkable success in generating images with a fixed resolution. However, existing models have limited ability to generalize to different resolutions when training data at those resolutions are not available. Leveraging techniques from operator learning, we present a novel deep-learning architecture, Dual-FNO UNet (DFU), which approximates the score operator by combining both spatial and spectral information at multiple resolutions. Comparisons of DFU to baselines demonstrate its scalability: 1) simultaneously training on multiple resolutions improves FID over training at any single fixed resolution; 2) DFU generalizes beyond its training resolutions, allowing for coherent, high-fidelity generation at higher-resolutions with the same model, i.e. zero-shot super-resolution image-generation;
    3) we propose a fine-tuning strategy to further enhance the zero-shot super-resolution image-generation capability of our model, leading to a FID of 11.3 at 1.66 times the maximum training resolution on FFHQ, which no other method can come close to achieving. 
\end{abstract}

\section{Introduction}

Diffusion models have been shown to be a powerful method for image generation \cite{sohl2015deep, SGM,Ho2020DenoisingDP, CascadedDiffusion, edm, dhariwal2021diffusion, Rombach2021HighResolutionIS}. Of particular interest is the diffusion model's ability to generate high-resolution images. One approach is to generate lower-resolution samples and then upscale using existing learning-based methods \citep[e.g.,][]{CascadedDiffusion, IterativeRefinement, gigagan}. Also notable is image outpainting \citep[e.g.,][]{dongConvolutionalSuperresolution, LDM}, which increases the resolution of an image by extending its content outside of the original palette of the diffusion model, again in a supervised fashion. However, both approaches require access to \textit{high-resolution training data} and do not truly generalize the learned score beyond training resolution. In contrast, we seek to learn a model which can directly \textbf{sample at high resolutions} to generate the entire image after being \textbf{trained at  low resolutions}, without access to high resolution training data. \footnote{Code: \url{https://github.com/Dahoas/edm}} 

We call this problem of generating resolution higher than that of training data \texttt{zero-shot super-resolution generation}. To tackle it, we start off by first considering a standard diffusion process as a discretization of an infinite dimensional generative process\cite{daprato_zabczyk_1992}. Under appropriate conditions we can recover the backward process with an associated score operator \cite{FOLLMER198659,InfiniteDimensionalDiffusion,DiffusionFunctionSpace}. Then we propose Dual-Fourier Neural Operator UNet (DFU) as a scale-robust deep learning architecture that can approximate, and in fact further improve the score operator across multiple resolutions. Evaluations on complex real-world image datasets demonstrate DFU's superior zero-shot super-resolution generation capabilities. We observe that it is able to generate samples up to twice the resolution of the training data with good \textit{coherence} (the overall global image structure) and \textit{fidelity} (the clarity of fine-details). It is also especially suitable for mixed-resolution training, which can significantly improve image quality at all resolutions when compared to single-resolution training. In particular, multi-resolution DFU achieves superior FID to even single resolution UNet models evaluated at their training resolution. Notably, this benefit is only enjoyed by DFU, as the same mixed-resolution training procedure seems to slightly harm UNet image quality.

\paragraph{Related Work} Operator learning via deep neural networks has exploded in popularity in recent years \cite{Lu2019LearningNO, NeuralOperator, Goswami2022PhysicsInformedDN, SMAI-JCM_2021__7__121_0}.  This work uses Fourier Neural Operator (FNO) \citep{Li2020FourierNO}, which learns solution operators to PDEs by composing blocks which simultaneously learn linear transformations in the spatial and Fourier domains. This architecture has led to many follow-up works \cite{GANO, Wen2021UFNOA,MultiwaveletPDE}. Notably, recent independent work by \citet{Rahman2022UNOUN} extends the FNO architecture to UNet for solving PDEs. 

Diffusion process in infinite dim. and its time reversal have been extensively studied \cite{FOLLMER198659, daprato_zabczyk_1992, Millet}. Recent works have also adopted this perspective with the purpose of learning diffusion generative models in infinite dimensions. However these papers focus more on constructing a rigorous formulation of such processes with empirical experiments learning Gaussian process distributions \cite{kerrigan2023diffusion}, well-behaved PDEs \cite{DiffusionFunctionSpace}, or simple image datasets such as MNIST \cite{hagemann2023multilevel}. In contrast, this paper focuses on the design and training of neural networks to approximate the score operator for significantly more complex distributions with less regularity. We also note concurrent work \citep{franzese2023continuoustime} which proposes \textit{Functional Diffusion Processes} to generate complex image distributions.

\section{Designing and training Dual-FNO UNet}
\paragraph{Learning the score operator} An infinite dimensional denoising diffusion process can be constructed if one has the corresponding score operator (see \ref{app:theory} for details or \cite{InfiniteDimensionalDiffusion,DiffusionFunctionSpace}). We wish to design an architecture that can approximate the score operator across \textit{multiple resolutions}. It was believed UNet \citep{unet}, a popular architecture used in the majority of diffusion models \citep{Ho2020DenoisingDP, SGM, Rombach2021HighResolutionIS}, is multiscale, and in fact, UNet does have shown its ability to learn high-frequency components of images via spatial convolutions at multiple scales. In addition, the field of operator learning offers explicit designs of neural networks capable of learning mappings between function spaces, such as for solving PDEs \cite{NeuralOperator,Li2020FourierNO, GANO, Rahman2022UNOUN}. 

However, as will be demonstrated in our experiments, both of these choices come with drawbacks for our purpose of zero-shot super-resolution generation. The fact that UNet relies on spatial convolutions makes it difficult to maintain global coherence (i.e. lower frequency information) when sampling beyond training resolution (see e.g., Fig. \ref{fig:conv-vs-dualfno}). On the other hand, FNO was designed to learn mappings between regular functions, e.g., those commonly appearing as strong solutions to PDEs. Meanwhile images are less smooth due to sharp edges and maintaining only lower-frequency information results in blurred images at higher resolutions.

\paragraph{Designing Dual-FNO UNet}
These observations lead us to introduce the Dual-Convolution ($DC$) of a function $f: \R^d \to \R$ as:
\begin{align}
DC(f) &=   \kspatial * f + \cF^{-1} ({\kspectral} \odot \cF(f) ) \in \cH
\label{eqdualconv}
\end{align} 
We call the first term the \textit{spatial convolution} and the second term the \textit{spectral convolution}. We use this operation as the base for DFU by introducing it into the Unet architecture. (For precise definitions refer to \ref{app:dual-fno-defs}). We implement DFU with $L=4$ ``UNet layers'' and 4 Dual-FNO blocks per layer. Each Dual-FNO block consists of 2 Dual-Convolutions with an affinely transformed time-embedding added in between. A spatial kernel size of $k=3$ is used at all levels. The $M=16$ lowest Fourier modes are used in the top level spectral convolution. The number of modes used is then halved at each subsequent level. A full diagram of the architecture is attached in the appendix \ref{fig:dualUnet}.

Note we tweaked the spatial convolution part of the implementation: the size of the discrete kernel remains fixed even if resolution increases, and thus it is not directly a discretization but a rescaled version. This implies that our architecture is not an operator. Note this is by design, as it allows the learning of fine scale details at lower resolutions, which then can be transferred to higher scale and become fine\textbf{r} scales. For instance, when trained on $96\times96$ data, the spectral convolution would enlarge a $2\times2$ patch of the image to $4\times 4$ when sampling a $192\times192$ image. However, the spatial convolution would remain $2\times2$ ensuring the details are refined at the smallest scale possible. This contributes to DFU's ability of zero-shot generation of higher resolution images with high fidelity, setting it apart from pre-existing  architectures or techniques such as FNO \cite{Li2020FourierNO}, UNO \cite{Rahman2022UNOUN}, or bilinear interpolation. 


\section{Experiments}
Code: \url{https://github.com/Dahoas/edm}

\begin{figure}
    \centering
    \includegraphics[scale=1]{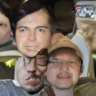}
    \includegraphics[scale=0.6]{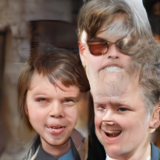}
    \includegraphics[scale=0.75]{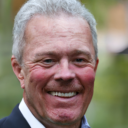}
   \caption{Visual comparison of DFU to various baselines at zero-shot super-resolution image generation. \textbf{Panel 1:} Single-resolution UNet trained on data with resolution up to 96x96, sampled at 128x128. \textbf{Panel 2:} Multi-resolution UNet sampled at 128x128. \textbf{Panel 3:} Multi-resolution DFU sampled at 128x128. Note both UNets struggle to generalize global coherency past the max training resolution $r=96$.
   }
   \label{fig:conv-vs-dualfno}
\end{figure}

\paragraph{Setup} We evaluate on Flicker Faces HQ (FFHQ) \citep{Karras2018ASG} and Lsun-Church \citep{Yu2015LSUNCO}. For each dataset, we start with a ground truth resolution $r=$ 256.
To construct our mixed-resolution training data, we downsample to resolutions $r \in \{32, 48, 64, 80, 96\}$. Training is run for 400k steps with a batch size of 256. Sampling is done in 18 steps using the EDM scheme \cite{edm}. Note, while the majority of discussion here analyzes FFHQ, a comparison of results for LSUN can be found in Appendix \ref{app:data-dist}.

\begin{wrapfigure}{R}{0.5\textwidth}
    \centering
    \includegraphics[scale=0.45]{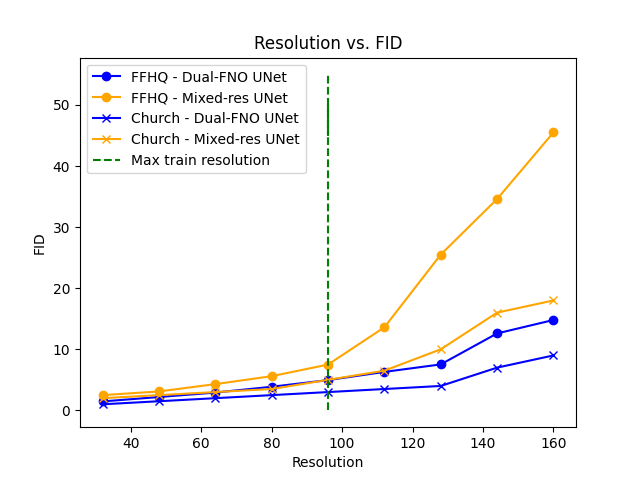}
   \caption{Sampled resolution versus FID of multi-res UNet and DFU. DFU has lower FID before losing local coherence at 2x training resolution.}
   \label{fig:res-fid}
\end{wrapfigure}

\begin{table}[ht]
\begin{center}
\begin{tabular}{ |p{0.3\linewidth}|c c c|c c| } 
 \hline
   & 32x32 & 64x64 & 96x96 & \textcolor{red}{128x128} & \textcolor{red}{160x160} \\ 
 \hline\hline
 Single-res UNet + Bilinear & 2.9 & 5.3 & 6.4 & \textcolor{red}{14.5} & \textcolor{red}{31.9} \\
 \hline
 Multi-res UNet & 2.5 & 4.3 & 7.5 & \textcolor{red}{25.2} & \textcolor{red}{45.5} \\ 
 \hline
  FNO UNet & 30.5 & 40.1 & 31.4 & \textcolor{red}{53.5} & \textcolor{red}{64.1} \\
 \hline
 DFU & \textbf{1.47} & \textbf{2.92} & \textbf{4.96} & \textcolor{red}{\textbf{7.85}} & \textcolor{red}{\textbf{14.8}} \\ 
 \hline
\end{tabular}
\end{center}
\caption{
FIDs on \textbf{FFHQ} across resolutions. Note all models are trained on a maximum resolution of $r=$ 96. Evaluation done on \textcolor{red}{$r= $ 128, 160} is \textcolor{red}{zero-shot super-resolution} image generation. }
\label{table:fid}
\end{table}
\paragraph{Baselines} We compare DFU against several baselines. \textit{Single-res UNet + Bilinear} is the standard \textit{single-resolution UNet} trained on a fixed resolution $r =$ 96. Bilinear interpolation is then used to resize an image to the target resolution. The \textit{Multi-resolution UNet} architecture adapts single-resolution UNet to training on multiple resolutions. We achieve this by removing the resolution dependent operations from a vanilla UNet. \textit{FNO UNet} replaces the spatial convolutions in DFU with kernels of size 1, mimicking the FNO architecture per block. This architecture is similar to the UNO architecture in \cite{Rahman2022UNOUN}. Notice that by comparing against these baselines we demonstrate the need for both spatial and spectral convolutions. Table \ref{table:fid} reports a subset of the results for FFHQ.  Other than for single-resolution Unet, which is trained on a fix resolution, training is run on the same mixture of resolutions as used to train DFU. 
Figure \ref{fig:res-fid} plots $\textup{FID}_r$ on both datasets for DFU and the best performing baseline. Full tables for all FIDs from both datasets are provided in the appendix.

\begin{figure}[!b]
    \centering
    \includegraphics[scale=0.8]{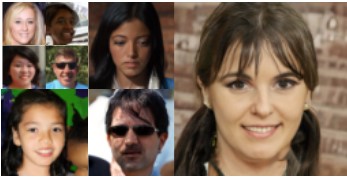}
    \includegraphics[scale=0.7]{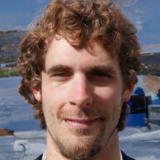}
   \caption{Samples from DFU across resolutions. \textbf{Left:} DFU sampled at $r=$ 32, 64, 128. \textbf{Right:} DFU sampled at $r=$ 160. DFU is trained on a mixture of resolutions from $r=$ 32 to 96.}
\label{fig:dualFNO-samples}
\end{figure}

\paragraph{DFU generalizes to higher resolutions} DFU achieves a lower FID score than all baselines across all training resolutions. Furthermore, DFU  significantly outperforms all baselines in the zero-shot super-resolution image-generation setting. The closest contender is the single-res UNet + bilinear interpolation baseline which can maintain coherent facial structures at higher resolutions via but quickly loses fidelity. For a direct comparison between the two approaches see Figure \ref{fig:bilinear-vs-dualunet}. In contrast both multi-res UNet and FNO UNet struggle, with multi-res UNet unable to learn structural coherency and FNO UNet unable to learn high-frequency details. For a direct comparison between super-resolved images from the multi-res UNet and our model, see Figure \ref{fig:conv-vs-dualfno}. Further we remark DFU is able to generalize, not just the low-frequency structure of images to higher resolutions, but also higher-frequency details than presented in the training data. See for example Figure \ref{fig:dualFNO-samples} which displays the generation ability of our model across multiple resolutions. These observations suggest spatial convolution learn detail while low-frequency spectral convolutions learn structure.  

\paragraph{Resolution training mixture impacts zero-shot super-resolution image-generation} By default we train all models on a uniform mixture of training resolutions. 
We also experiment with adjusting this mixture to place higher probability on higher resolution samples. However, we found concentrating too much mass on high resolutions leads to overfitting. This damages zero-shot super-resolution performance. Our best performing mixture sets resolution weight $w_\textup{96} = 0.4$ and $w_\textup{80} = 0.3$ with the rest of the weight decaying exponentially. The resulting model achieves slightly better FID on higher training resolutions and worse FID on lower resolutions than uniform mixing. However, we see a significant improvement in zero-shot super-resolution image-generation FID, e.g., improving $\textup{FID}_{\textup{160}}$ on FFHQ to 12.1 from the 14.8 of our uniformly mixed model. 

\begin{table}[ht]
\begin{center}
\begin{tabular}{ |p{0.3\linewidth}|c c c|c c c| } 
 \hline
  Model & 32x32 & 64x64 & 96x96 & \textcolor{red}{128x128} & \textcolor{red}{160x160}  & \textcolor{red}{196x196}\\ 
 \hline\hline
 Dual FNO Unet & \textbf{1.31} & 2.40 & 3.43 & \textcolor{red}{\textbf{4.86}} & \textcolor{red}{\textbf{8.78}} & \textcolor{red}{\textbf{20.00}} \\
 \hline
 Multi-res Unet & 1.52 & \textbf{2.29} & 3.62 & \textcolor{red}{6.67} & \textcolor{red}{11.95} & \textcolor{red}{20.69} \\ 
 \hline
 Single res Unet & 127.28 & 27.94 & \textbf{3.31} & \textcolor{red}{15.44} & \textcolor{red}{32.06} & \textcolor{red}{NA} \\
 \hline
 Single res Unet + Bilinear Interpolation & 2.96 & 3.26 & 3.31 & \textcolor{red}{7.15} & \textcolor{red}{15.08} & \textcolor{red}{23.71} \\
 \hline
\end{tabular}
\end{center}
\caption{
FIDs on \textbf{LSUN Church} across resolutions for different models. The single res Unet model is trained on $96\times96$ data.}
\label{table:fids-church}
\end{table}

\paragraph{Mixed-resolution training improves single-resolution generation}
Surprisingly, DFU significantly outperforms the baselines, not just in the zero-shot super-resolution generation regime, but also within training resolutions. For example, on FFHQ our model attains $\textup{FID}_{\textup{96}} = 4.96$ in comparison to a vanilla UNet architecture with $\textup{FID}_{\textup{96}} = 6.4$. This suggests mixed-resolution training with DFU can improve $\textup{FID}_r$ over training only at the single resolution $r$. To test this hypothesis, we train a series of single-resolution UNet and DFU models from $r = 32$ to $r=96$. We find DFU, when trained at a single resolution, slightly under-performs UNet. However, when trained at multiple resolutions, DFU outperforms single resolution UNet models. This even holds for the lowest resolutions, where UNet trained on $r=$32 achieves $\textup{FID}_{\textup{32}} = 1.75$ whereas our $\textup{FID}_{\textup{32}} = 1.47$. Full tables can be found in the appendix.  




\bibliographystyle{unsrtnat}
\bibliography{arxiv.bib}

\section*{Acknowledgment}

We sincerely thank Tianrong Chen, Guan-Horng Liu and Qinsheng Zhang for precious tips, such as for training existing models. AH is grateful to Stability.AI for partial support from a Graduate Research Fellowship and providing compute to train the models. KR and MT are grateful for partial support from NSF DMS-1847802, GT Cullen-Peck Scholarship, and GT-Emory Humanity.AI Award. 

\appendix

\section{Diffusion and Diffusion Generative Modeling in Infinite Dimensions}
\label{app:theory}
We first review the approach in \citep{InfiniteDimensionalDiffusion} for defining an infinite dim. forward SDE, its time-reversal, and the score operator: 
Let $\mathcal{D}$ be a data distribution supported on a separable Hilbert space with inner product $\langle \cdot ,\cdot \rangle$. For a fixed positive-definite, symmetric covariance operator $C: \cH\rightarrow \cH$ with ${\rm trace}(C) <\infty$, the forward SDE is:
$$dX_t = -\frac 1 2 X_t dt +\sqrt{C} dW_t, \quad X_0\sim \mathcal{D}$$
where $C W_t$ is called the C-Wiener process \citep{daprato_zabczyk_1992}.
As $t\rightarrow +\infty$, the distribution of $X_t$ will converge to the stationary distribution $\mathcal{N}(0,C)$. 

The score, generalized to infinite dim as an operator, can be defined as \citep{InfiniteDimensionalDiffusion}
\begin{equation}
     s(t,x) = \frac{1}{1- e^{-t}}\left(x - e^{\frac{-t}{2}} \E[X_0|X_t = x]\right), \ \text{ for } x\in \cH, 
     \label{eqscoreoperator}
\end{equation}
and it will correct the drift in the infinite dim. backward SDE
\begin{equation*}
dY_t = \frac 1 2 Y_t dt + s(t,Y_t)dt +\sqrt{C} dW_t, \quad Y_0  \sim p_T
\end{equation*}
so that again $Y_t$ and $X_{T-t}$ agree in distribution.

\begin{remark}
An alternative approach to infinite dim diffusion generative model is adopted in \cite{hagemann2023multilevel}, where one can define the process using a countable orthonormal basis (thanks to separable Hilbert space) $\{e_i\}_{i = 1}^\infty$ where each $e_i$ is an eigenvector of $C$ and their associated eigenvalues are $\lambda_1 \geq \lambda_2 \geq \lambda_3 \geq \dots \geq 0$. One can then define the process of interest coordinate-wise \citep{FOLLMER198659} and discretize as
\begin{align}\label{eq:inf-dim-diff}
    &dX_t^i = f(X_t^i,t) dt + g(t) dW_t^{C_n}, \qquad X_0 \sim D \qquad i = 1,\dots, n
\end{align} 
where $W_t^C$ denotes $\sqrt{C} W_t$ and $C_n$ is a $n$-th truncated Karhunen-Lo\`eve decomposition of $W_t^C$, i.e: 
\[W_t^{C_n} = \sum_{k = 1}^n \sqrt{\lambda_k} \beta_k(t) e_k\]

for mutually independendent standard Wiener processes $\beta_k$. These processes can then be treated as standard finite dimensional processes. However, we move away from this approach, since we would like to parametrize the score as an operator.
\end{remark}

\begin{center}
    \begin{figure}[t]
    \hspace{-2.5cm}
    \includegraphics[width=1.4\linewidth]{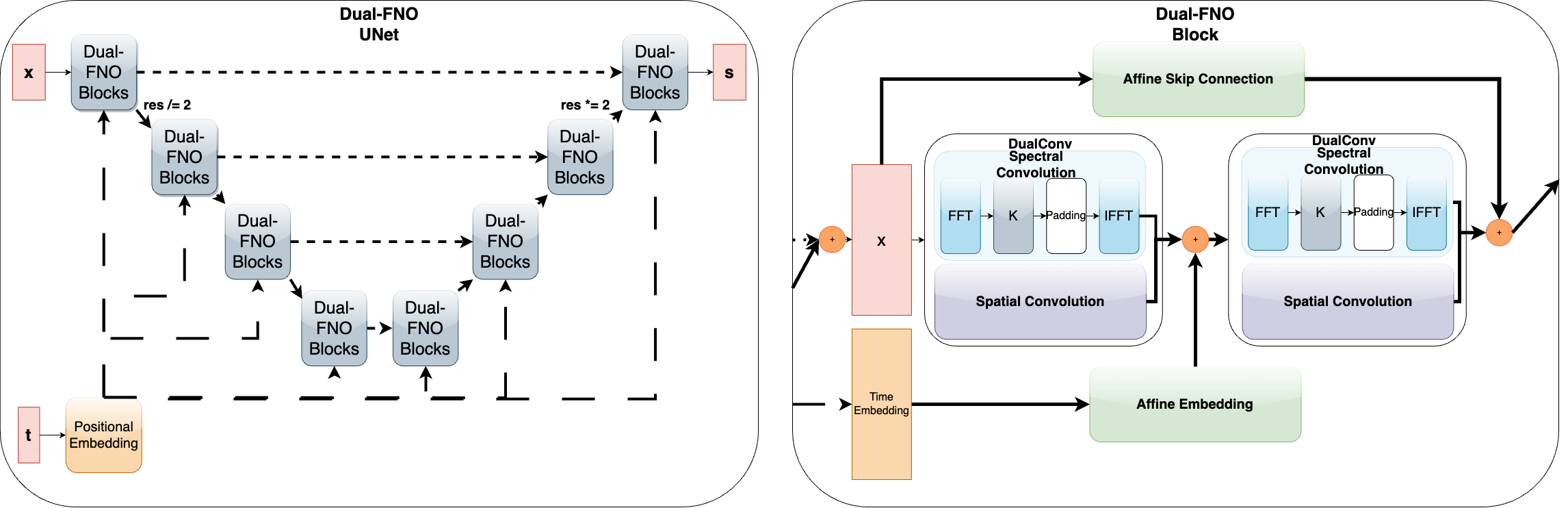}
   \caption{\textbf{Left:} DFU architecture. \textbf{Right:} Dual-FNO Block. Right is integrated in left by connecting arrows of corresponding line styles (solid for passing matrices, short dashed for skip connection, and long dashed for time embedding).}
   \label{fig:dualUnet}
\end{figure}
\end{center}

\section{Dual FNO Precise Definitions}
\label{app:dual-fno-defs}
\paragraph{Images as discretizations of functions} We model 2D images as functions on $[0,1]^2$. In particular, we consider 2D image with three channels to correspond to a spatial discretization of three functions of continuous 2D space in the separable Hilbert space $\cH = L^2([0,1]^2)$. An image at resolution $r$ discretizes an image function $f$ by sampling $f$ from a uniform grid of $r^2$ points $G_r \subseteq [0,1]^2$. The key to achieving zero-shot super-resolution image generation is then to design a model which can learn features of the underlying image function distribution from a set of training image resolutions. Note for ease of presentation our further description will often be based on operations on functions from 1D space $[0,1]$ to $\mathbb{R}$. However, generalizing to 2D space (and higher-dim. if needed) and multiple channels is natural and simply done coordinate-wise.

\paragraph{FNO Blocks}  FNO is composed of a series of \textit{FNO blocks} of the form 
\begin{align*}
    B_{\textup{FNO}}(f) = W \odot f + \mathcal{F}^{-1}(K \odot \mathcal{F}(f))
\end{align*} where $\mathcal{F}(f)$ is the Fourier transform of $f$, and $\odot$ is Hadamard (i.e. pointwise) product. 
$W, K\in\cH$ are two learnable functions, and $K$ induces a linear operator $f\mapsto \mathcal{F}^{-1}(K \odot \mathcal{F}(f))$ that is a pointwise multiplication in frequency domain but a convolution in spatial domain. 






In practice, $f$ and $K$ are both discretized. In order for the learned $K$ to extend to different resolutions, FNO adopts the approach of fixing its bandwidth, i.e. keep the number of nonzero elements in $K$ constant based on a fixed cutoff frequency.

\paragraph{Dual Convolution}
We introduce the Dual-Convolution ($DC$) of a function $f: \R^d \to \R$ as:
\begin{align}
DC(f) &=   \kspatial * f + \cF^{-1} ({\kspectral} \odot \cF(f) ) \in \cH
\label{eqdualconv}
\end{align} where * is convolution and $\kspatial: \mathbb{R}^d \rightarrow \R$ is a function supported on $[-\varepsilon_s,\varepsilon_s]^d$, corresponding to a learnable 
spatial kernel, and $\kspectral: \R^d \rightarrow \C$ is a function supported on $[-M,M]^{d}$ (i.e. with cut-off frequency $M>0$), corresponding to a learnable spectral kernel. We call the first term the \textit{spatial convolution} and the second term the \textit{spectral convolution}. 

Note that these two components are not equivalent since $\kspatial$ is localized in space and $\kspectral$ is localized in frequency. Due to the Heisenberg Uncertainty Principle \citep{heisenberg1985anschaulichen}, a kernel cannot be localized in space and frequency simultaneously. Thus the spatial kernel $\kspatial$ cannot be replaced by $\cF^{-1} (\kspectral)$ or vice versa. The \emph{DC} operator  augments the analagous spectral convolution in FNO such that the global low-frequency features can be learned by the spectral convolution and the high-frequency features can be learned by the spatial  convolution.

\section{Fine-tuning via bilinear Interpolation}
\begin{figure}[ht]
    \centering
    \includegraphics[scale=0.6]{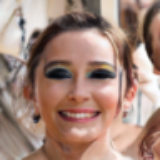}
    \includegraphics[scale=0.6]{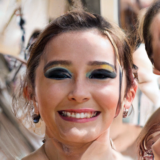}
    \includegraphics[scale=0.6]{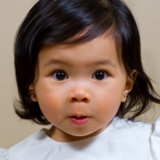}
   \caption{Comparison of pre-trained DFU to bilinear upsampling. \textbf{Left:} $r=$96 image bilinearly upsampled to 160x160. \textbf{Middle:} DFU sampled at 160x160. \textbf{Right:} Ground truth 160x160 image included as a reference for quality. The bilinearly upsampled image is able to maintain good coherency but lacks fidelity. In contrast DFU maintains both coherence and fidelity comparable to the ground truth.}
   \label{fig:bilinear-vs-dualunet}
\end{figure}

Despite improved resolution generalization compared to baselines, DFU still suffers from coherency issues when sampled at a sufficiently high resolution. However, image fidelity remains largely unaffected. 
In contrast, simple zero-shot super-resolution generation techniques such as bilinear interpolation maintain image coherency at the cost of greatly reduced fidelity. For example, see the left two panels of Figure \ref{fig:bilinear-vs-dualunet} and compare the images sampled via DFU and bilinear interpolation.

\paragraph{Setup} To improve DFU for large resolutions, we leverage bilinear interpolation's ability to maintain coherence when upsampling. To do so, we first upsample our maximum resolution training dataset $\mathcal{D}_{r_{\texttt{max}}}$ to a target resolution $R > r_{\texttt{max}}$ producing our fine-tuning dataset $\mathcal{D}_{R}$. We then take our model pre-trained on lower resolutions and fine-tune on $\mathcal{D}_{R}$ to improve image coherence at resolution $R$. Note this bilinear interpolation itself is a zero-shot super-resolution technique. Therefore our model is not being exposed to ground-truth higher-resolution images.

While fine-tuning, we need to do so without destroying the superior fidelity learned by DFU. To preserve fidelity, we use two independent tricks. First, we fine-tune on a mixture of resolutions instead of just on $\mathcal{D}_R$. We assign mix weight $w_R$ to $\mathcal{D}_R$ and the the rest of the mix weight uniformly assigned to the pre-training resolutions as $w_{r} = \frac{1 - w_R}{5}$ for $r \in \{32, 48, 64, 80, 96\}$. This prevents our model from overfitting to the poor fidelity of $\mathcal{D}_R$ by simultaneously re-training on high fidelity low-resolution images. We also experiment with freezing the spatial convolutions of the pre-trained model when fine-tuning. This is motivated by the use of spatial convolutions to learn high-frequency details (providing good fidelity) with the unfrozen spectral convolutions learning low-frequency structure to improve coherency.

\begin{table}[ht]
\begin{center}
\begin{tabular}{ |p{0.2\linewidth}|c c c|c c| } 
 \hline
  & 32x32 & 64x64 & 96x96 & 128x128 & 160x160 \\ 
 \hline\hline
 Pre-trained Model & 1.47 & 2.92 & 4.96 & 7.85 & 14.8 \\ 
 \hline
 Vanilla fine-tune ($w_R = 1.0$) & 20.92 & 12.94 & 14.96 & 32.21 & 27.1 \\ 
 \hline
 Mix w/ $w_R = 0.2$ & 1.41 & \textbf{2.44} & 4.66 & 7.69 & 12.4 \\ 
 \hline
 Frozen spatial conv & 1.46 & 2.85 & 4.76 & 7.62 & 11.9 \\
 \hline
 Frozen spatial conv + $w_R = 0.2$ & \textbf{1.36} & 2.46 & \textbf{4.54} & \textbf{7.06} & \textbf{11.3} \\
 \hline
\end{tabular}
\end{center}
\caption{
FIDs on FFHQ across resolutions for different fine-tuning schemes. Our best result combines mixed resolution fine-tuning and parameter freezing.}
\label{table:finetuned-fids}
\end{table}

\paragraph{Results} We sweep over the mixing parameter $w_R$, the learning rate, and the number of fine-tuning steps while additionally freezing/un-freezing parts of the architecture. Results are recorded in table \ref{table:finetuned-fids}. Our best results are achieved with $w_R = 0.2$, $lr = 1e-4$, while fine-tuning for an additional 10k steps with all the spatial convolutions frozen except those on the bottom layer of DFU. 


We find both mixed resolution fine-tuning with a small $w_R$ and freezing spatial convolutions to be important for preserving image fidelity at higher resolutions. The choice of $w_R$ naturally trades-off how much fidelity is learned from lower resolutions versus how much coherence is learned from the upsampled dataset $\mathcal{D}_R$. Higher $w_R$ corresponding to improved coherence. Beyond a certain threshold (e.g., $w_R \geq 0.5$) image fidelity degrades to a quality similar to $\mathcal{D}_R$.  We see a full fine-tune with $w_R = 1$ degrades lower resolution performance. Additionally, fine-tuning for too many steps (more than 15k) allows the model to overfit to the poor fidelity data in $\mathcal{D}_{\textup{256}}$. This adversely trades-off quality for coherence.

Freezing the spatial convolution blocks in combination with mixed resolution training further improves image fidelity allowing us to obtain our best zero-shot 11.3 FID on $r=$160. When fine-tuning on a mixed resolution dataset with frozen spatial conv blocks we found it best to unfreeze the model on lower resolution samples, only freezing parameters on batches from $\mathcal{D}_R$. Fine-tuning with frozen blocks without a mixed dataset improved fidelity over the baseline vanilla fine-tune, but not as much as with mixed-resolution training. See Figure \ref{fig:finetune-fidelity} for a comparison of various methods.

Fine-tuning on low-quality high resolution data also has the added benefit of further improving FID on lower resolutions. For example this improves our $r=$32 FID to a remarkable 1.36, despite fine-tuning on data at 5 times this resolution. 
This suggests learning structural details to improve coherence at higher resolutions allows our model to improve its score approximation across all resolutions. Such improvement gives more evidence that DFU well approximates the underlying score operator.

\begin{figure}[ht]
    \centering
    \includegraphics[scale=0.6]{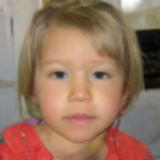}
    \includegraphics[scale=0.6]{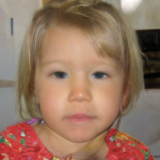}
    \includegraphics[scale=0.6]{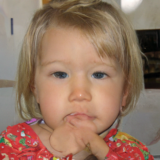}
   \caption{A visual comparison between fine-tuning methods. \textbf{Left:} 160x160 image sampled from naively-finetuned model. The image is over-smoothed. \textbf{Middle:} 160x160 image sampled from frozen + mixed resolution fine-tuned model. We have sacrificed some fidelity (but not too much) for coherency. \textbf{Right:} 160x160 image sampled from pre-trained model (i.e. DFU trained on resolution up to 96x96, without fine-tuning). This image has good fidelity but poor coherence.}
   \label{fig:finetune-fidelity}
\end{figure}

\section{FID across resolutions}

In this section we add all the FID values for different models across resolutions for both FFHQ and LSUN-Church.
\label{app:fid-tables}

\begin{table}[ht]
\begin{center}
\addtolength{\leftskip} {-2cm}
\addtolength{\rightskip}{-2cm}
\begin{tabular}{ |p{0.3\linewidth}|c c c c c|c c c c| } 
 \hline
  & 32x32 & 48x48 & 64x64 & 80x80 & 96x96 & \textcolor{red}{112x112} & \textcolor{red}{128x128} & \textcolor{red}{144x144} & \textcolor{red}{160x160} \\ 
 \hline\hline
 Single-res UNet & 1.75 & 2.65 & 3.92 & 4.85 & 6.43 & \textcolor{red}{N/A} & \textcolor{red}{N/A} & \textcolor{red}{N/A} & \textcolor{red}{N/A} \\
 \hline
 Single-res Dual-FNO UNet & 2.05 & 3.21 & 4.38 & 5.46 & 7.11 & \textcolor{red}{N/A} & \textcolor{red}{N/A} & \textcolor{red}{N/A} & \textcolor{red}{N/A} \\
 \hline
 Single-res UNet (96x96) + Bilinear & 2.9 & 4.4 & 5.3 & 5.5 & 6.4 & \textcolor{red}{11.6} & \textcolor{red}{14.5} & \textcolor{red}{23.2} & \textcolor{red}{31.9} \\
 \hline
 Multi-res UNet & 2.5 & 3.6 & 4.3 & 5.8 & 7.5 & \textcolor{red}{17.9} & \textcolor{red}{25.2} & \textcolor{red}{36.4} & \textcolor{red}{45.5} \\
 \hline
 FNO UNet & 12.1 & 17.4 & 27.4 & 29.1 & 31.4 & \textcolor{red}{36.8} & \textcolor{red}{44.5} & \textcolor{red}{48.9} & \textcolor{red}{52.1} \\
 \hline
 Uniformly Pre-trained Dual-FNO UNet & 1.47 & 2.24 & 2.92 & 3.86 & 4.96 & \textcolor{red}{6.43} & \textcolor{red}{7.85} & \textcolor{red}{11.6} & \textcolor{red}{14.8} \\ 
 \hline
 Weighted Pre-trained Dual-FNO UNet & 2.56 & 3.55 & 3.94 & 3.72 & 4.80 & \textcolor{red}{6.01} & \textcolor{red}{7.12} & \textcolor{red}{10.3} & \textcolor{red}{12.1} \\
 \hline
 Vanilla fine-tune ($w_R = 1.0$) & 20.92 & 16.43 & 12.94 & 
13.37 & 14.96 & \textcolor{red}{21.52} & \textcolor{red}{32.21} & \textcolor{red}{35.2} & \textcolor{red}{27.1} \\ 
 \hline
 Mix w/ $w_R = 0.2$ & 1.41 & 1.94 & \textbf{2.44} & 3.54 & 4.66 & \textcolor{red}{6.31} & \textcolor{red}{7.69} & \textcolor{red}{10.2} & \textcolor{red}{12.4} \\ 
 \hline
 Frozen spatial conv & 1.46 & 2.45 & 2.85 & 3.77 & 4.76 & \textcolor{red}{5.95} & \textcolor{red}{7.62} & \textcolor{red}{9.92} & \textcolor{red}{11.9} \\
 \hline
 Frozen spatial conv + $w_R = 0.2$ & \textbf{1.36} & \textbf{1.89} &  2.46 & \textbf{3.48} & \textbf{4.54} & \textcolor{red}{\textbf{5.87}} & \textcolor{red}{\textbf{7.06}} & \textcolor{red}{\textbf{9.77}} & \textcolor{red}{\textbf{11.3}} \\
 \hline
\end{tabular}
\end{center}
\caption{
FIDs on \textbf{FFHQ} across resolutions for all model types. Note the single-res UNet/Dual-FNO UNet rows train and evaluates models on each resolution independently.Evaluation done at more than $96\times 96$ is  \textcolor{red}{super-resolution}.}
\label{table:finetuned-fids}
\end{table}

\begin{table}[ht]
\begin{center}
\begin{tabular}{ |p{0.3\linewidth}|c c c|c c c| } 
 \hline
  Model & 32x32 & 64x64 & 96x96 & \textcolor{red}{128x128} & \textcolor{red}{160x160}  & \textcolor{red}{196x196}\\ 
 \hline\hline
 Dual FNO Unet & \textbf{1.31} & 2.40 & 3.43 & \textcolor{red}{\textbf{4.86}} & \textcolor{red}{\textbf{8.78}} & \textcolor{red}{\textbf{20.00}} \\
 \hline
 Multi-res Unet & 1.52 & \textbf{2.29} & 3.62 & \textcolor{red}{6.67} & \textcolor{red}{11.95} & \textcolor{red}{20.69} \\ 
 \hline
 Single res Unet & 127.28 & 27.94 & \textbf{3.31} & \textcolor{red}{15.44} & \textcolor{red}{32.06} & \textcolor{red}{NA} \\
 \hline
 Single res Unet + Bilinear Interpolation & 2.96 & 3.26 & 3.31 & \textcolor{red}{7.15} & \textcolor{red}{15.08} & \textcolor{red}{23.71} \\
 \hline
\end{tabular}
\end{center}
\caption{
FIDs on \textbf{LSUN Church} across resolutions for different models. The single res Unet model is trained on $96\times96$ data, the other models are trained on a uniform mixture of $32\times32$, $48\times48$, $64\times 64$ and $96\times 96$ data. Evaluation done at more than $96\times 96$ is  \textcolor{red}{super-resolution}.}
\label{table:fids-church}
\end{table}




\section{Dependence on Data Distribution}
\label{app:data-dist}
We observe better results on the LSUN-Church dataset compared to FFHQ. We attribute this difference to the relatively more rigid global structure of FFHQ. LSUN-Church contains a broader spectrum of potential global structures, as churches can assume various forms. Additionally, the level of detail surrounding the churches is more lenient, allowing for greater flexibility. Conversely, the FFHQ dataset is less flexible, as the global structure of a face must be preserved consistently, and the details must closely resemble the actual features (eyes, noses, etc.). Thus distributions with rigid global structure like FFHQ hihglight the ability of DFU to well learn this structure. A full table of FIDs for both FFHQ and LSUN-Church can be found in the appendix along with samples from LSUN-Church.

\section{LSUN-Church samples across resolutions}
\label{app:lsun-samples}
In this section we provide additional  samples generated on the LSUN-Church dataset using Dual FNO Unet. We train our model using the same configuration as for the FFHQ dataset and the same multi-resolution training regime. No fine tuning was performed for this dataset.

\begin{center}
 \begin{figure}[ht]
    \centering
    \includegraphics[scale=0.6]{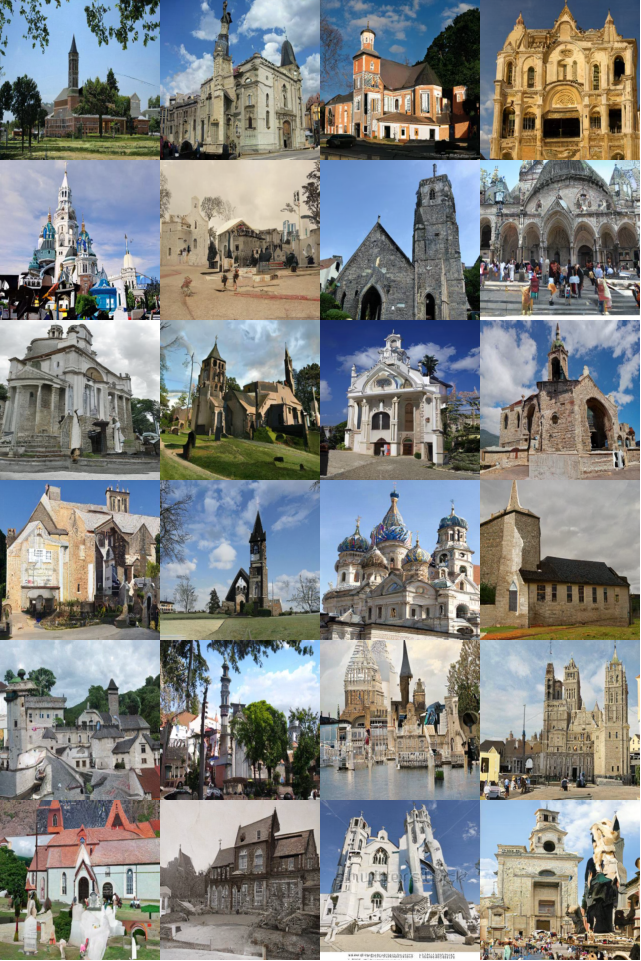}
   \caption{Samples from Dual FNO Unet at $160\times160$ resolution. The model was trained in resolutions up to $96\times96$}
   \label{fig:finetune-fidelity}
\end{figure}   
\end{center}

  \begin{figure}[ht]
    \hspace{-1cm}
    \includegraphics[scale=0.6]{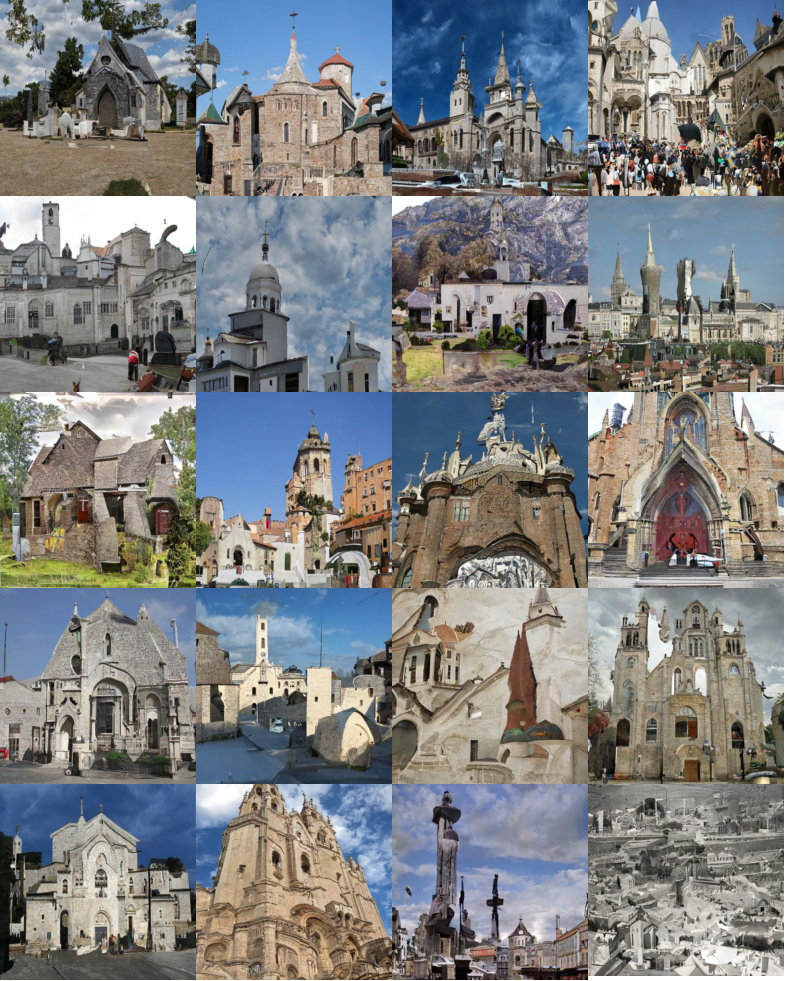}
   \caption{Samples from Dual FNO Unet at $196\times196$ resolution. The model was trained in resolutions up to $96\times96$}
   \label{fig:finetune-fidelity}
\end{figure}  

\begin{figure}[ht]
\hspace{-1cm}
    \includegraphics[scale=0.6]{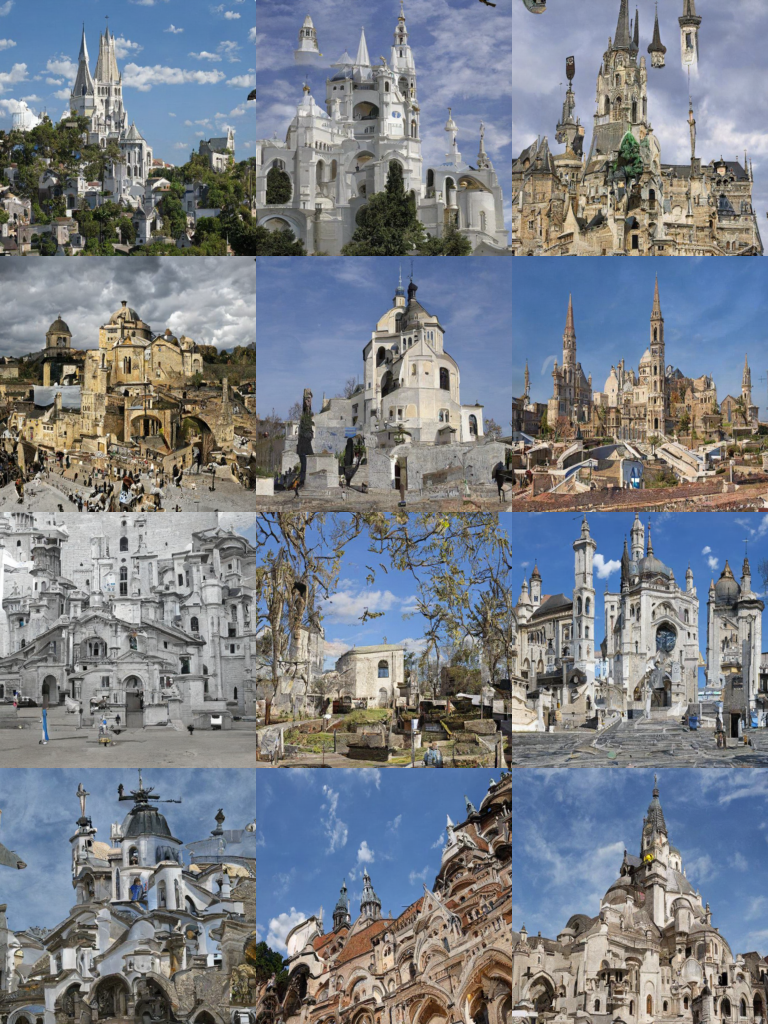}
   \caption{Samples from Dual FNO Unet at $256\times256$ resolution. The model was trained in resolutions up to $96\times96$}
   \label{fig:finetune-fidelity}
\end{figure}

\section{FFHQ samples across resolutions}

\begin{figure}[ht]
\hspace{-1cm}
\begin{center}
    \includegraphics[scale=0.7]{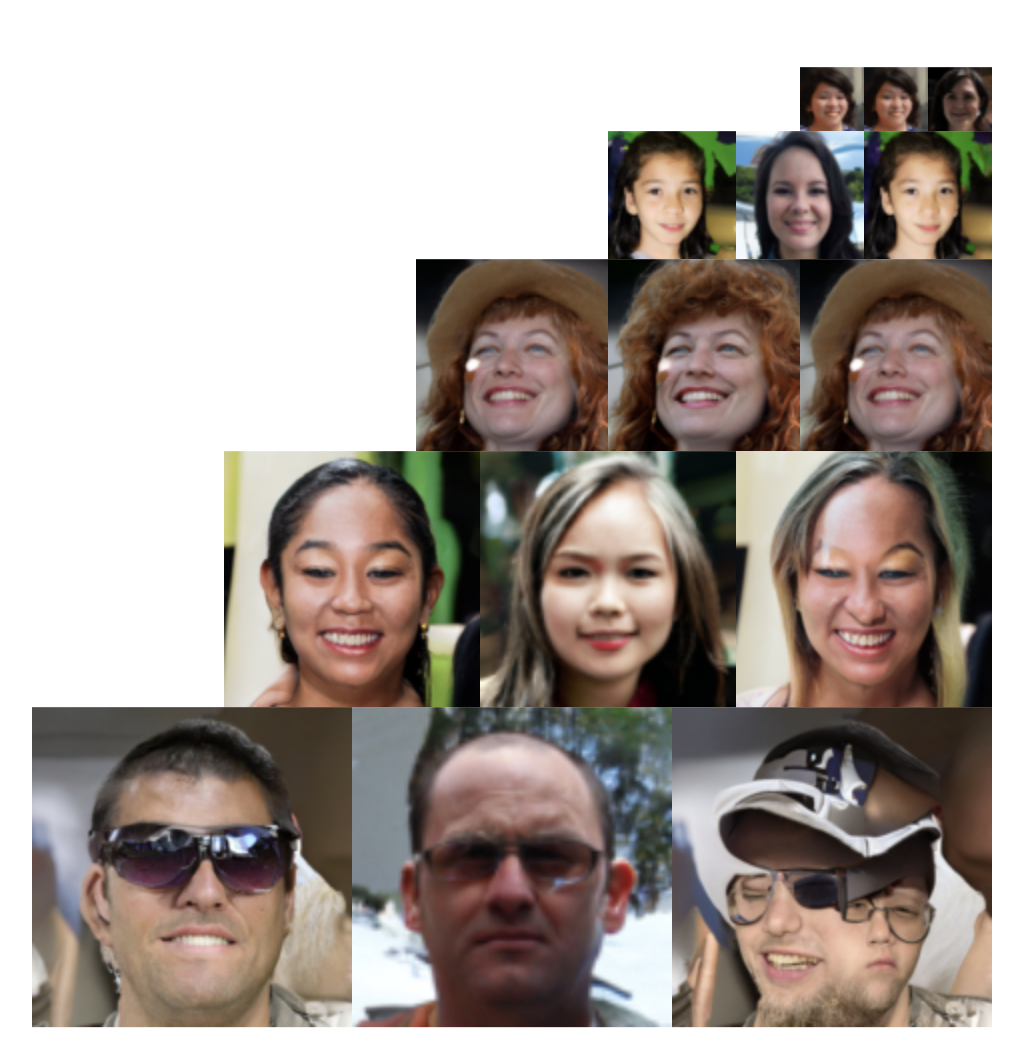}
\end{center}
   \caption{Samples from DFU, SingleRes + Bilinear Interpolation, and MultiRes UNet respectively up to $160\times160$ resolution. The models were trained in resolutions up to $96\times96$}
   \label{fig:ffhq-compare}
\end{figure}

\begin{figure}[ht]
\hspace{-1cm}
    \includegraphics[scale=0.7]{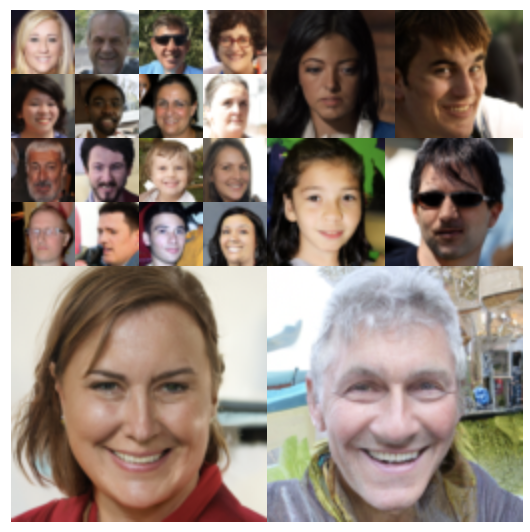}
    \includegraphics[scale=0.7]{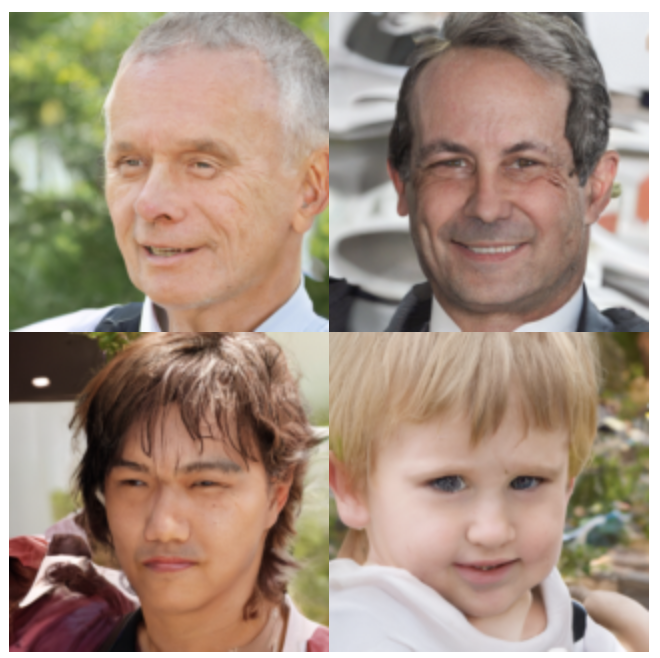}
   \caption{\textbf{Left:} Samples from DFU over $32\times 32$ to $128\times 128$. \textbf{Right:} DFU at $160 \times 160$}
   \label{fig:ffhq-dfu}
\end{figure}

\begin{figure}[ht]
\hspace{-1cm}
    \includegraphics[scale=0.7]{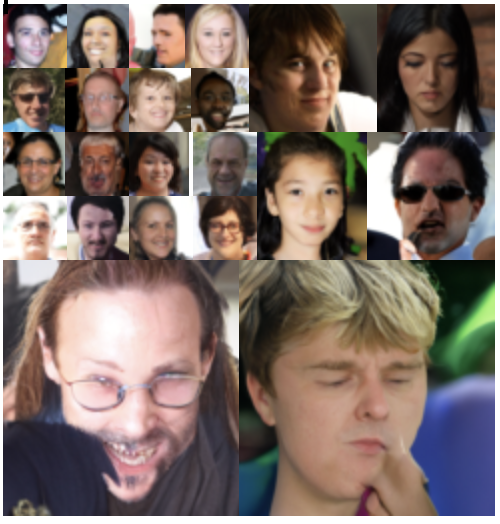}
    \includegraphics[scale=0.7]{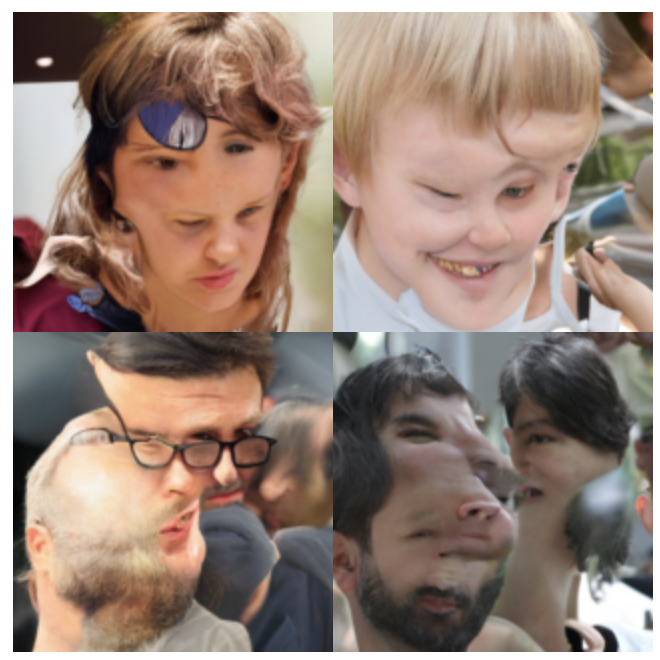}
   \caption{\textbf{Left:} Samples from MultiRes UNet over $32\times 32$ to $128\times 128$. \textbf{Right:} MultiRes UNet at $160 \times 160$}
   \label{fig:ffhq-mix-res}
\end{figure}

\begin{figure}[ht]
\hspace{-1cm}
    \includegraphics[scale=0.7]{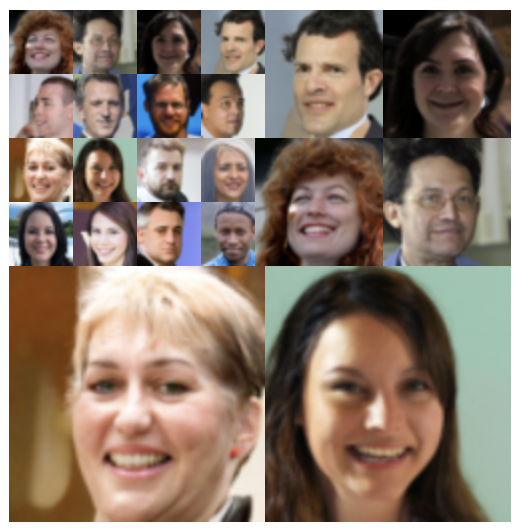}
    \includegraphics[scale=0.7]{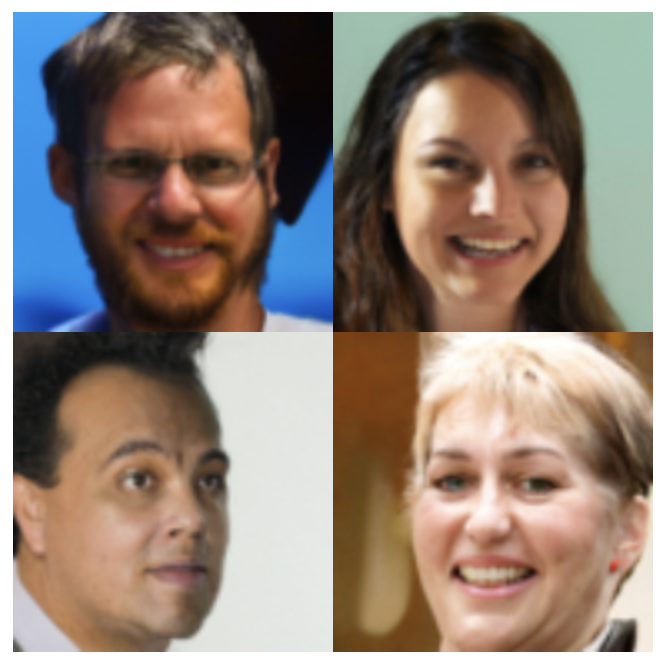}
   \caption{\textbf{Left:} Samples from SingleRes UNet + bilinear interpolation over $32\times 32$ to $128\times 128$. \textbf{Right:} SingleRes UNet + bilinear interpolation at $160 \times 160$}
   \label{fig:ffhq-single-res}
\end{figure}

We provide additional sampled generated by DFU, the MultiRes baseline, and the SingleRes baseline on FFHQ. Figure \ref{fig:ffhq-compare} provides a direct visual comparison of the methods. Figures \ref{fig:ffhq-dfu}, \ref{fig:ffhq-mix-res}, and \ref{fig:ffhq-single-res} contain more samples from each method respectively.

\end{document}